\pdfoutput=1
\documentclass[11pt]{article}

\usepackage{ACL2023}

\usepackage{times}
\usepackage{latexsym}
\usepackage{graphicx}

\usepackage[T1]{fontenc}
\usepackage[utf8]{inputenc}

\usepackage{microtype}

\usepackage{inconsolata}
\usepackage{amsmath}
\usepackage{amssymb}
\usepackage{multirow}
\usepackage{booktabs}
\usepackage{pifont}
\usepackage{url}
\usepackage{tablefootnote}

\makeatletter
\newcommand\footnoteref[1]{\protected@xdef\@thefnmark{\ref{#1}}\@footnotemark}
\makeatother

\title{Universal-2-TF: Robust All-Neural Text Formatting for ASR}

\author{Yash Khare\thanks{$^*$Authors contributed equally.} , Taufiquzzaman Peyash$^*$, Andrea Vanzo, Takuya Yoshioka \\ 
        AssemblyAI Inc. \\
        \url{https://www.assemblyai.com/research/universal-2}}

\begin{document}
\maketitle
\begin{abstract}
This paper introduces an all-neural text formatting (TF) model designed for commercial automatic speech recognition (ASR) systems, encompassing punctuation restoration (PR), truecasing, and inverse text normalization (ITN). Unlike traditional rule-based or hybrid approaches, this method leverages a two-stage neural architecture comprising a multi-objective token classifier and a sequence-to-sequence (seq2seq) model. This design minimizes computational costs and reduces hallucinations while ensuring flexibility and robustness across diverse linguistic entities and text domains. Developed as part of the Universal-2 ASR system, the proposed method demonstrates superior performance in TF accuracy, computational efficiency, and perceptual quality, as validated through comprehensive evaluations using both objective and subjective methods. This work underscores the importance of holistic TF models in enhancing ASR usability in practical settings.
\end{abstract}

\section{Introduction}
\label{sec:intro}

Automatic speech recognition (ASR) systems often produce output text in spoken form, requiring text formatting (TF) post-processing to convert the ASR model's output into a written style. This conversion enhances the readability of the generated transcripts and improves compatibility with various downstream processes. While some ASR models--particularly sequence-to-sequence (seq2seq) models like Whisper \cite{radford2023robust}—-are trained on audio files with written-form transcripts collected from the Internet and can directly generate properly formatted transcripts, separating ASR into speech-to-text (STT)\footnote{For clarity, this paper uses ``STT'' to refer to the process of converting speech audio into spoken-form transcripts, while ``ASR'' refers to generating written-form transcripts, although these terms are often used interchangeably.} and TF offers practical advantages, making TF remain essential in modern commercial ASR systems. It enables consistent enforcement of desired formats for representing various linguistic entities and allows for the effective utilization of large amounts of text-only data to train TF models. Meanwhile, STT can focus exclusively on spoken content, which is particularly beneficial for reducing word error rates (WERs), especially in low-resource languages, and enables the use of relatively small STT models.

In commercial settings, TF typically requires multiple functionalities, including punctuation restoration (PR), truecasing, and inverse text normalization (ITN). PR adds punctuation marks, such as periods and commas, at appropriate locations within a given word sequence. Truecasing determines the correct casing of individual letters, such as capitalizing the first word of a sentence, proper nouns, and acronyms (e.g., representing acronyms in all caps) while handling mixed-case words, such as McDonald's and JavaScript. ITN converts spoken-form entities into their written equivalents, including but not limited to ordinals, currencies, dates, postal addresses, email addresses, URLs, and social security numbers (SSNs). Combining these functionalities would transform spoken-form text, such as \texttt{``on fifteenth march two thousand and twenty-four ceo sarah mcallister announced that aicorps's revenue reached twelve point three million dollars''}, into a polished written form: \texttt{``On March 15th, 2024, CEO Sarah McAllister announced that AICorp’s Q1 revenue reached \$12.3 million.''}

Conventionally, processes such as PR, truecasing, and ITN have been studied independently, leaving the question of how best to combine these individual components unresolved. Additionally, many existing methods, particularly for ITN, rely on rule-based approaches using weighted finite-state transducers (WFSTs)~\cite{DBLP:journals/corr/abs-2104-05055} or a hybrid of neural and rule-based techniques~\cite{pusateri2017mostly,9414912,10022543}. While these approaches are computationally efficient and less prone to hallucinations, they are limited in accuracy and generalizability in handling a broad range of ITN entities.

This paper presents a fully-fledged \textit{all-neural} TF method that performs PR, truecasing, and ITN. The method was developed as part of AssemblyAI's Universal-2, a state-of-the-art commercial ASR system, which has been demonstrated to outperform other open-source and commercial ASR systems\footnote{\url{https://www.assemblyai.com/research/universal-2}}. The method comprises two neural network models that work together to perform the aforementioned tasks. The first model is a multi-objective token classifier that handles PR and capitalization while identifying textual spans that may require ITN or mixed-casing. The multi-objective token classifier has a shared encoder and multiple heads, which is demonstrated to reduce inference cost without accuracy loss. The second model is a seq2seq model applied to the identified spans to perform ITN and mixed-case word conversion. By limiting the text segments' length processed by the seq2seq model, the proposed method achieves a practically affordable computational cost and avoids catastrophic hallucinations while benefiting from the enhanced flexibility provided by the seq2seq model compared to WFSTs. Delegating the handling of mixed-case words to the seq2seq model, in addition to ITN, eliminates the need for error-prone character-level casing decisions. 
Comprehensive evaluation results are presented, covering quantitative metrics for TF accuracy, computational cost, and perceptual quality, demonstrating the effectiveness of the proposed all-neural method as well as highlighting the need for schemes that holistically evaluate various aspects of TF.

\section{Related Work}
\label{sec:rw}

TF for ASR is essentially a sequence conversion task from normalized text, or a sequence of uncased spoken words, to a written format. With the prevalence of Transformers, it might be tempting to use a seq2seq Transformer model with a bidirectional encoder and an autoregressive decoder to achieve this. However, this approach suffers from practical challenges, such as high computational costs required for producing long text as well as a lack of robustness to hallucinations, which are critical in large-scale usage scenarios. In fact, to date, many ASR systems built for large-scale deployment have adopted a hybrid approach combining multiple modules, each designed for solving subtasks such as PR, truecasing, and ITN, as we review in this section.

\paragraph{Punctuation Restoration (PR)}
PR has often been cast as a sequence labeling task, where each token or word constituting an input sequence is labeled with an appropriate punctuation symbol, such as a period and a comma, as well as a blank symbol representing no punctuation.
\citet{tilk2015lstm} pioneered the sequence labeling approach by employing a two-stage LSTM. 
More recent work used pre-trained encoders to better capture long-term dependencies \cite{courtland2020efficient,guerreiro2021towards}.
Some previously proposed models attempted to solve additional tasks, such as disfluencies processing \cite{lin2020joint} and truecasing \cite{nguyen2019fast}. It is also noteworthy that utilizing acoustic or prosody input in addition to textual observations was demonstrated to be helpful \cite{zelasko2018punctuation,zhu2024resolving}. 

\paragraph{Truecasing}
Unlike PR, truecasing has been mostly cast as a character classification task to deal with mixed-case words, such as McDonald's, JavaScript, and AssemblyAI. RNNs \cite{susanto2016learning,ramena2020efficient} and Transformer-based hierarchical approaches \cite{zhang2022capitalization} were often used in prior studies. 
However, in practice, character-based approaches suffer from the \textit{``not all errors are created equal''} problem. That is, showing ``JavaScript'' as ``JavAScrIpt'' is much more detrimental to perceived quality than ``javascript'' while the error count is equal between these two examples. Therefore, some guardrail measures are often required to deploy such models, which could nullify the truecasing capability.
While several studies adopted word-level approaches~\cite{nguyen2019fast,sunkara-etal-2020-robust,9687976}, they typically simplified the truecasing task to an easier capitalization task by ignoring or only partially covering mixed-case words.

\paragraph{Inverse Text Normalization (ITN)}
ITN, a common challenge in real-world ASR, is a task of converting spoken entities in a transcript into their written form. ITN must handle a wide range of entity classes, including ordinals, currencies, dates, postal addresses, email addresses, URLs, and SSNs, among others.

Conventionally, rule-based approaches using handcrafted WFST grammars~\cite{DBLP:journals/corr/abs-2104-05055} were employed for ITN. However, these approaches are cumbersome and expensive to scale across a wide range of linguistic entities and languages, and have limited ability to leverage linguistic context. 

More recently, neural approaches have been explored, offering enhanced capabilities for utilizing context and improved generalization to scenarios not explicitly accounted for during system design.
Most methods are based on sequence labeling:  an input normalized text is processed with a Transformer encoder followed by classification heads to label each token, where the labels specify how to convert the corresponding input tokens to a written format. Such conversion can be achieved by using WFSTs~\cite{pusateri2017mostly,9414912,10022543} or other handcrafted rules~\cite{paul2022improving,antonova2022thutmose}. However, reliance on rule-based approaches in the final step still limits the generalization and scalability of these models. 

Notably, \citet{nguyen2023adapitn} proposed combining encoder-only and encoder-decoder Transformers, where the labels predicted by the encoder-only Transformer are used to identify input text spans to be processed with the additional Transformer decoder. This method leverages the Transformer decoder's ability to model any type of conversion learned during training. In addition, it is computationally efficient since the decoder is applied only to limited textual spans, addressing the challenge of seq2seq model mentioned at the beginning of Section \ref{sec:rw}. 

\paragraph{End-to-end Text Formatting (TF)}
Unlike the methods reviewed above, which address only specific subtasks of TF, 
\citet{tan2023four} presented a fully-fledged TF method that handles all the aforementioned subtasks, except for processing mixed-case words. The method employs a multi-head classification model that jointly performs labeling tasks for punctuation restoration, capitalization, and ITN pre-processing, followed by WFST-based text conversion.

\hspace*{.5em}

Drawing inspiration from \citet{nguyen2023adapitn} and \citet{tan2023four}, we propose an all-neural TF model. Our model leverages a Transformer decoder, or a seq2seq model, to handle various entities requiring ITN without relying on handcrafted conversion rules. The seq2seq model is applied only to specific textual spans, which helps maintain inference costs within a practically affordable range and minimizes the risk of hallucinations, eliminating the need for additional guardrail measures. These textual spans are identified by a multi-head classifier that simultaneously performs PR and capitalization. Notably, our seq2seq model also handles mixed-case words, enabling the entire system to achieve truecasing, unlike existing methods.

\section{Proposed Model: Universal-2-TF}
\label{sec:model}

The proposed TF model, named \textit{Universal-2-TF}, is based on a pipeline comprising two models, a multi-objective token classifier and a seq2seq model. Both models are based solely on text generated by STT, with no acoustics utilized.

\begin{figure*}
    \centering
    \includegraphics[width=\textwidth]{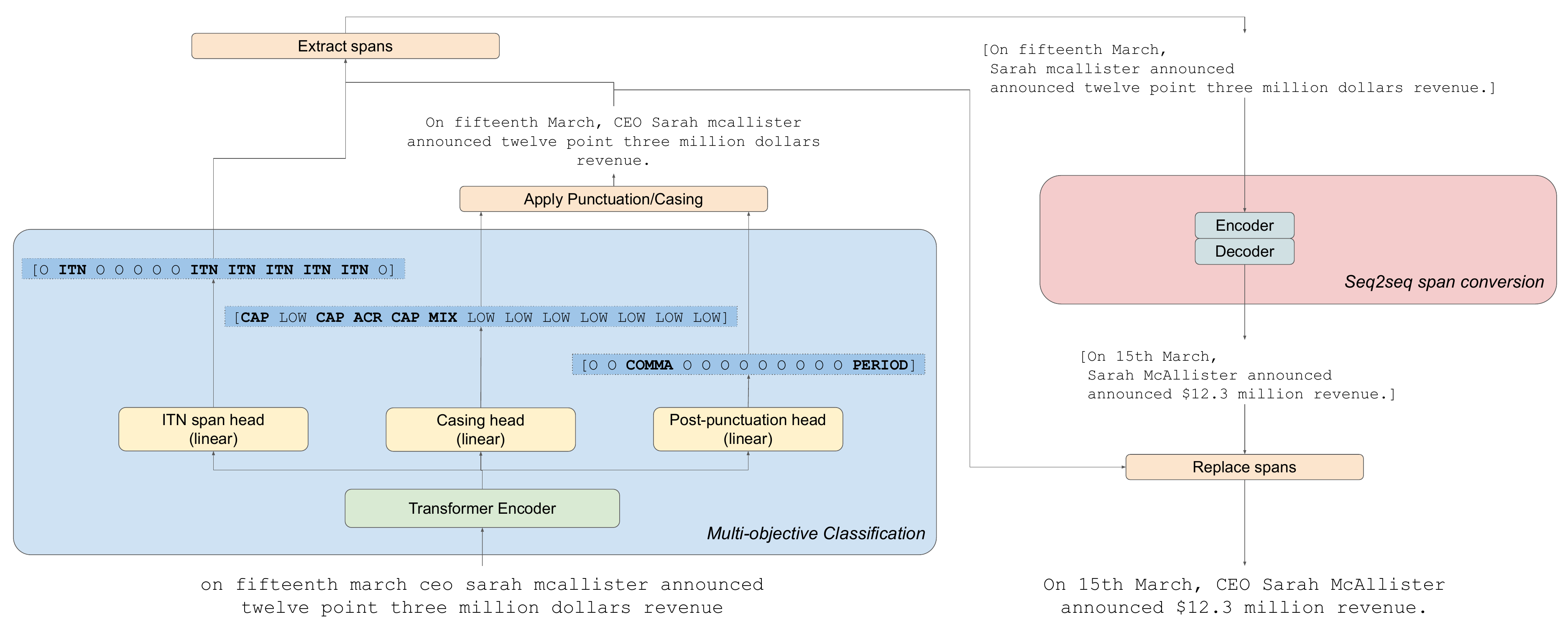}
    \caption{Universal-2-TF model architecture: A Transformer-based encoder generates token representations of the input text, which are processed by punctuation, truecasing, and ITN span identification heads. Punctuation and truecasing predictions are applied to the input text, from which ITN spans and mixed-case words are extracted along with limited left and right context (one word in this diagram). The identified spans are then processed by a seq2seq model for conversion and reintegrated into the original text.}
    \label{fig:architecture}
\end{figure*}

Figure~\ref{fig:architecture} shows the overall model architecture. The first stage performs multi-objective token classification using a multi-head model, predicting punctuation marks and token-level casing labels while identifying textual spans requiring mixed casing and ITN. The second stage uses a seq2seq encoder-decoder model to convert short, unformatted textual spans identified in the previous stage into their formatted expressions to jointly perform mixed-casing and ITN. The model utilizes some left and right context for improved accuracy. The predictions obtained from both stages are applied to the text. 

The proposed architecture offers several key advantages, as discussed below. First, the use of a shared encoder in the multi-objective classification model enables efficient inference by jointly performing PR, truecasing, and span detection tasks, while potentially capturing the correlations inherent in these tasks.

Second, a seq2seq architecture is crucial for handling ITN and mixed-case word conversion, as these tasks cannot be easily modeled as classification problems. The seq2seq model generates text autoregressively, enabling more flexible text conversion while leveraging both left and right contexts through a bidirectional Transformer encoder.

Third, avoiding full transcript processing provides two key benefits. First, focusing on restricted textual spans minimizes computational overhead and reduces processing time. Second, this approach allows the model to be fine-tuned for specific ITN and mixed-case conversion tasks, enhancing its effectiveness without sacrificing robustness in handling generic text conversion cases.

Finally, delegating the mixed-case transformation to the seq2seq model offers practical benefits for truecasing. This approach allows us to use a token-level method in the first stage, as other truecasing transformations, such as sentence capitalization and all-caps conversion, do not require per-character decisions. This results in a reduction in computational cost compared to using a character-based truecasing model.

The remainder of this section describes the two models in detail. 

\subsection{Multi-Objective Token Classification}
\label{sec:backbone}

The first model is a multi-objective network undertaking multiple token classification tasks. Given an input text, a sequence of feature representations is first obtained through a Transformer-based encoder. Those representations are then fed to three \textit{linear} heads to predict token-level labels, with each head focusing on different tasks through disjoint label sets, namely punctuation restoration, truecasing, and ITN span identification.

Formally, given an input token sequence $X = \{x_1, \ldots, x_n\}$ provided by an STT model, each classification head predicts a label sequence of the same length:
\begin{align*}
    L^k &= \{l_1^k, l_2^k, \ldots, l_n^k\}, (k = 1, 2, 3)
\end{align*}
where:
\begin{itemize}
    \item $l_i^1\in\{\mathtt{PERIOD},\mathtt{COMMA},\mathtt{QUESTION},\mathtt{O}\}$ denotes a post-punctuation mark to be appended to the $i$-th input token, with $\mathtt{O}$ representing no punctuation; 
    \item $l_i^2\in\{\mathtt{CAPITAL},\mathtt{ACRONYM},\mathtt{MIXED}, \mathtt{LOWER}\}$ denotes a  truecasing label for the $i$-th token, with $\mathtt{MIXED}$ used to identify mixed-case words to be processed with the subsequent seq2seq network;
    \item $l_i^3\in\{\mathtt{ITN},\mathtt{O}\}$ denotes whether the $i$-th token is part of a  word to be inverse-normalized. 
\end{itemize}

We train the three-headed model, including both the encoder and the task-specific classification heads, by minimizing a combined loss function that aggregates cross-entropy losses $\mathcal{L}_k$ from each classification head:
\begin{equation}
\mathcal{L} = \alpha_1\mathcal{L}_1 + \alpha_2\mathcal{L}_2 + \alpha_3\mathcal{L}_3
\end{equation}
with $\alpha_k$ being pre-determined task-specific weights such that $\sum_{k=1}^3 \alpha_k = 1$ and $\alpha_k > 0$. In our experiments, we set $\alpha_k = 1/3$ for all $k$ values for simplicity.

At inference time, the predicted labels from each head are used as follows. 
\begin{itemize}
    \item \textit{Punctuation restoration}: When the predicted label is any of $\{\mathtt{PERIOD}, \mathtt{COMMA}, \mathtt{QUESTION}\}$, the input token is appended with the corresponding punctuation mark. 
    \item \textit{Truecasing}: When the predicted label is either $\mathtt{CAPITAL}$, the first letter of the input token is capitalized whereas if the predicted label is $\mathtt{ACRONYM}$, all letters  constituting the input token are capitalized. When the predicted label is  $\mathtt{MIXED}$, a text span surrounding the current token, including some left and right contexts, is extracted for further text transformation using the seq2seq model. 
    \item \textit{ITN span identification}: Neighboring tokens with predicted labels of $\mathtt{ITN}$ are grouped together, along with some left and right contexts, to form ITN text spans. Individual text spans are then processed by the seq2seq model for ITN. 
\end{itemize}
Note that predictions for subword tokens are ignored. 

\subsection{Seq2seq Text Span Conversion}
\label{sec:seq2seq}

The second stage is a seq2seq language model based on an architecture using a bi-directional encoder and an autoregressive decoder \cite{lewis-etal-2020-bart}. It is trained to perform ITN and convert lowercase words into properly mixed-cased expressions. 

Given a sequence of tokens in each span identified from the first stage, $X = \{x_1, ..., x_n\}$, 
the model predicts an output token sequence $Y = \{y_1, ..., y_m\}$.
This is achieved by finding the output token sequence that maximizes a posterior probability $P(Y|X) = P(y_1, ..., y_m | x_1, ..., x_n)$, where the probability distribution is learned during training. 
The bi-directional encoder of the seq2seq model computes hidden states $H = \{h_1, ..., h_n\}$, which are used to condition the decoding process. The decoder generates the output sequence $Y$ auto-regressively using a greedy search strategy, that is, the output tokens are produced one at a time. At each time step $t$, the decoder finds the token that maximizes the posterior probability over the token vocabulary, 
\begin{equation}
P(y_t | y_{<t}, H), 
\end{equation}
where $y_{<t}$ represents the sequence of previously generated tokens.

The model is trained by using a dataset consisting of input-output textual pairs. The training is achieved by minimizing a cross-entropy loss.

\section{Data Processing Pipeline}
\label{sec:data}

Our goal is to learn a mapping from normalized text to formatted text using the multi-objective classification and seq2seq models described in the previous section. The training data for these models can be generated from a textual corpus by applying text normalization techniques. This can be achieved by using the Nvidia NeMo Text Normalizer~\cite{DBLP:journals/corr/abs-2104-05055}, which leverages WFSTs.

Although conceptually straightforward, achieving highly accurate and robust text formatting involves addressing several intricacies in data processing. This section describes these details.

\subsection{Quality First: Data Cleaning and Filtering}
\label{sec:processing}

Data cleaning and filtering are crucial to ensure that models are exposed to high-quality, diverse datasets during training. Publicly sourced data often contain unwanted artifacts that could negatively impact specific tasks. Also, such data may not align well with our target application domains. To address these issues, we developed a generic pre-processing pipeline and tailored it to generate training data separately for the multi-objective classification and seq2seq models. 

More specifically, our data preparation pipeline consists of three steps:
\begin{enumerate}
    \item \textit{Coarse-grained filtering}: We discard data points that deviate significantly from typical punctuation usage and capitalization patterns.
    \item \textit{Text cleaning and normalization}: We preprocess the text by removing out-of-scope symbols, correcting known errors and capitalization inconsistencies, and eliminating transcript artifacts, such as speaker labels, from text data sourced from transcribed spoken audio.
    \item \textit{Fine-grained filtering}: We further refine the dataset by applying stricter filtering criteria to obtain the cleaned text.
\end{enumerate}
Details of the data cleaning process are provided in the Appendix.

The pipeline is used to generate data for both the multi-objective token classifier and the seq2seq model. We adjust its configurations to optimize for the specific needs of each model.
Early empirical investigations revealed that the multi-objective model benefits more from data diversity and scale than from data quality, whereas the seq2seq model is highly sensitive to training data quality and requires higher quality datasets. 
Also, we used tailored cleaning and normalization schemes to account for the different vocabularies required by each model. For example, currency symbols are essential for the seq2seq model to correctly predict corresponding linguistic entities, while they are not expected for the multi-objective model, which operates exclusively on normalized text.

\subsection{Filling the Void: Data Augmentation with LLMs}
\label{sec:synthetic}

One challenge in seq2seq model training is the under-representation of text formats with high practical importance in data from typical sources. Examples include proper nouns, acronyms, and entities requiring strict adherence to specific formatting guidelines, such as credit card numbers, phone numbers, and email addresses.

To address this, in addition to the data cleaning pipeline explained above, we expanded the seq2seq model's training data with synthetic textual examples generated by popular LLMs with permissive licenses. Specifically, we prompted the LLMs to generate diverse text samples rich in proper nouns, acronyms, and various linguistic constructs, as well as data containing entities that require specific text formatting schemes, like credit card numbers, phone numbers, and SSNs. Using these synthetic texts, generated by multiple LLMs,  provides the model with the ability to handle a wide range of practically important text formatting cases.

\section{Evaluation Results}
\label{sec:exp_setup}

We conducted experiments to evaluate the proposed Universal-2-TF model across various performance dimensions. 

The model used in our evaluation was constructed as follows:
\begin{itemize}
\item \textit{Multi-objective token classification model (first-stage model):} The encoder of this model was initialized with pre-trained BERT weights \cite{devlin-etal-2018-bert} using \textit{bert-base-uncased}\footnote{\url{https://huggingface.co/google-bert/bert-base-uncased}}. We then added randomly initialized linear heads on top of the encoder. All parameters were updated during fine-tuning. The model contained approximately 110M parameters.
\item \textit{Seq2seq model (second-stage model):} The second-stage model was initialized with a BART family model \cite{lewis-etal-2020-bart}. We specifically used \textit{bart-base}\footnote{\url{https://huggingface.co/facebook/bart-base}}, which achieved a good trade-off between accuracy and processing latency. During fine-tuning, all model parameters were updated. The model contained approximately 139M parameters.
\end{itemize}

\begin{table*}[]
\footnotesize
\centering
 \caption{Summary of training datasets consisting of a combination of public,\tablefootnote{\url{https://dumps.wikimedia.org}} purchased,\tablefootnote{\url{https://www.corpusdata.org}} in-house, and synthetic  data.}
 \vspace{-.5em}
 \label{tab:train_datasets}
\begin{tabular}{llrccc}
\toprule
 & \multicolumn{1}{r}{} & \multicolumn{1}{c}{\textbf{Word count}} & \multicolumn{1}{c}{\begin{tabular}[c]{@{}c@{}}\textbf{Avg. word count} \\ \textbf{per sample}\end{tabular}} & \multicolumn{1}{c}{\begin{tabular}[c]{@{}c@{}}\textbf{Used for 1st} \\ \textbf{stage training}\end{tabular}} &  \multicolumn{1}{c}{\begin{tabular}[c]{@{}c@{}}\textbf{Used for 2nd} \\ \textbf{stage training}\end{tabular}} \\
 \midrule
 \multirow{1}{*}{\textbf{Public}} & Wikipedia & 298M & 112.7 & & \checkmark  \\
 \midrule
 \multirow{6}{*}[0.6em]{\textbf{Purchased}} & CorpusData/NOW & 2,612M & 165.4 & & \checkmark  \\
 & CorpusData/SOAP & 77M & 133.9 & \checkmark & \checkmark  \\
 & CorpusData/COCA & 739M & 155.1 & \checkmark & \checkmark  \\
 & CorpusData/Movies & 152M & 141.8 & \checkmark & \checkmark  \\
 & CorpusData/TV & 252M & 144.3 & \checkmark & \checkmark  \\
 \midrule
 \multirow{2}{*}{\textbf{In-house}} & Human-labels & 1,949M & 150.0 & \checkmark & \checkmark  \\
 & Pseudolabels & 2,127M & 172.9 & \checkmark &  \\
 \midrule
 \textbf{Synthetic} & & 2,027M & 138.0 & & \checkmark  \\
 \midrule
 \multicolumn{2}{c}{\textbf{Total}} & 10.2B & & \multicolumn{1}{l}{} & \multicolumn{1}{l}{} \\
 \bottomrule
 \end{tabular}
 \end{table*}

Text data consisting of 10.2B words was used to fine-tune these models. Our training data comprised public, purchased, in-house, and synthetic datasets. See Table~\ref{tab:train_datasets} for details. The multi-objective token classification model was fine-tuned for 100k steps using a batch size of 256, utilizing all data related to the involved tasks (PR, truecasing, and ITN span prediction) as shown in the table. The seq2seq model was fine-tuned in two steps. First, we performed \textit{generic} fine-tuning on a larger, non-specialized dataset consisting of approximately 6B words, using the public and purchased datasets along with a subset of the in-house datasets. The model was trained for 500k iterations using a batch size of 512. In the second step, we further fine-tuned the model for 2k steps using a \textit{ITN-focused} dataset consisting of 2B words,  specifically generated through simulation for ITN and mixed-case word conversion. Both models were trained on v5e TPU clusters.

\subsection{Objective Evaluation Results}
To evaluate the entire TF system, we utilized five datasets to ensure variability in data distributions: SummScreen~\cite{chen2022summscreendatasetabstractivescreenplay}, DialogSum~\cite{chen2021dialogsumreallifescenariodialogue}, AMI~\cite{AMICorpus}, MeetingBank~\cite{hu-etal-2023-meetingbank}, and Europarl~\cite{koehn-2005-europarl}. The following performance metrics were employed. 
\begin{itemize}
\item \textit{Punctuation Error Rate (PER):} Measures prediction errors for commas, periods, and question marks, following the definition in \cite{meister2023librispeechpcbenchmarkevaluationpunctuation}.
\item \textit{ITN Word Error Rate (I-WER):} Quantifies errors specific to ITN and is calculated as follows. First, we remove all punctuation marks from both predicted and reference texts and convert them to lowercase. Then, we identify words requiring ITN by aligning the original, formatted reference text with its normalized version. Next, the formatted reference text is aligned with the predicted text, while ignoring punctuation and casing. 
We then calculate the WER over the words requiring ITN. 
\item \textit{Character Error Rate (CER):} Measures capitalization accuracy by computing CER after removing punctuation marks, following prior work~\cite{meister2023librispeechpcbenchmarkevaluationpunctuation}.
\item \textit{Mixed-case WER (M-WER):} Similar to I-WER, but WER is computed over mixed-case words in the reference transcription rather than words requiring ITN.
\item \textit{Inference time:} Measured on two datasets: short texts (average word count: 416) and long texts (average word count: 5,478). Processing time is recorded on an Nvidia T4 GPU.
\end{itemize}

\begin{table*}
\footnotesize
\centering
\caption{End-to-end evaluation results for punctuation, truecasing, and ITN accuracy, along with inference time. Punctuation accuracy is measured by PER, truecasing accuracy by CER and M-WER, and ITN accuracy by I-WER. Processing time results are shown for both short and long texts, separated by a forward slash (/).}
\label{tab:e2e_results}
\vspace{-.5em}
\begin{tabular}{lccccc}
\toprule
Model & PER (\%) & CER (\%) & M-WER (\%) & I-WER (\%) & Time (s)\\
\midrule
Universal-2-TF & \textbf{29.0} & \textbf{0.9} & \textbf{0.4} & \textbf{30.3} & \textbf{10.7}/\textbf{92.7}\\
~~~- w/o param. sharing& 29.0 & \textbf{0.9} & \textbf{0.4} & 30.4 & 12.7/130.0\\
Full seq2seq & 35.0 & 2.5 & 2.3 & 37.6 & 222.9/2845.8\\
Universal-1-TF & 29.9 & 1.2 & 0.6 & 52.7 & 11.2/127.9\\
\bottomrule
\end{tabular}
\end{table*}

Table~\ref{tab:e2e_results} shows the experimental results.
The row labeled ``w/o parameter sharing'' represents a model similar to Universal-2-TF, where first-stage token classification was performed using separate token classification models, each fine-tuned for individual tasks. These individual models were initialized and fine-tuned in the same manner as Universal-2-TF's multi-head model, resulting in an increase in the total parameter count of the entire system. The results show that Universal-2-TF achieved comparable accuracy across all performance dimensions while significantly reducing inference time. This demonstrates the effectiveness of sharing the encoder across all classification tasks in the first stage.

The table also compares Universal-2-TF with two internal models that perform PR, truecasing, and ITN. We selected these two models because no open-source system provides all the functionalities required for TF. The first model, \textit{full seq2seq}, is a seq2seq model based on a Transformer encoder/decoder, which directly processes normalized input text and generates formatted text. Its architecture is identical to that of the second-stage seq2seq model in Universal-2-TF and was trained on the same data used for training the Universal-2-TF seq2seq model. The result indicates that applying a seq2seq model in an end-to-end fashion not only leads to significant inefficiency in inference  but also results in degraded TF performance due to lack of robustness and hallucination. The second model, \textit{Universal-1-TF}, is our previous-generation TF system. It combines a neural PR model (similar to the first-stage model in Universal-2-TF), a character-based truecasing model, and a WFST-based ITN model, as described in \cite{assemblyai2024punctuation,ramirez2024anatomyindustrialscalemultilingual}. As expected, Universal-1-TF model exhibits significantly worse I-WER than Universal-2-TF, highlighting the limitations of WFST-based methods. Moreover, Universal-2-TF demonstrates superior performance in all other aspects, underscoring the robustness and efficiency of our proposed approach.

\subsection{Subjective Evaluation Results}

In addition to the objective evaluation described above, we conducted a subjective evaluation using human judges. While objective metrics such as PER and I-WER provide insights into model performance from large-scale experiments, they may not capture the nuanced qualities that humans appreciate in transcripts. In our subjective evaluation, we compared the proposed Universal-2-TF model with our previous TF method, Universal-1-TF, using 400 test samples. These test samples were carefully selected to represent the diversity of our commercial domains and ITN entities. To eliminate bias in quality judgments, we collaborated with two external vendors to recruit human judges, assigning six judges to evaluate each test sample. For each sample, formatted transcripts were generated using both models and presented side by side in a random order, with the models' identities concealed. The judges were then asked to indicate their preference for one of the models or provide a ``neutral'' rating.

\begin{table}[t]
\centering
\caption{Human preference scores between Universal-2-TF (proposed model), Universal-1-TF (previous model), and neutral.}
\label{tab:subjective_eval}
\vspace{-.5em}
\begin{tabular}{ccc}
\toprule
Universal-2-TF & Universal-1-TF & Neutral \\
\midrule
81.2\%        & 17.2\%        & 1.6\%   \\
\bottomrule
\end{tabular}
\end{table}

Table \ref{tab:subjective_eval} shows the win rates of the two models and the neutral judgment. Human judges expressed strong preferences to Universal-2-TF, demonstrating its superiority in perceived quality. 

\subsection{In-Depth Performance Analysis}

In addition to evaluating the proposed model in an end-to-end fashion, we investigated its performance across different TF categories using open-source models and previously reported results.

\begin{table*}[t]
\footnotesize
\centering
\caption{Punctuation F1 scores across different models and datasets.}
\label{tab:punctuation_results}
\vspace{-.5em}
\begin{tabular}{llccc}
\toprule
\multirow{2}{*}{\textbf{Dataset}} & \multirow{2}{*}{\textbf{Model}} & \multicolumn{3}{c}{\textbf{F1 score}} \\ \cmidrule(lr){3-5}

 & & $\mathtt{PERIOD}$ & $\mathtt{COMMA}$ & $\mathtt{QUESTION}$ \\
\midrule
\multirow{3}{*}{SummScreen} & Universal-2-TF & 83.8 & 74.8 & 83.8 \\
 & BadCode & 56.5 & 55.1 & 56.0 \\
 & Deep-Multilingual-Punct & 69.4 & 60.2 & 53.2 \\
\midrule
\multirow{3}{*}{DialogSum} & Universal-2-TF & 88.4 & 71.1 & 93.5 \\
 & BadCode & 71.3 & 60.0 & 76.7 \\
 & Deep-Multilingual-Punct & 81.2 & 66.0 & 85.3 \\
\midrule
\multirow{3}{*}{MeetingBank} & Universal-2-TF & 76.9 & 68.2 & 73.7 \\
 & BadCode & 58.9 & 57.0 & 55.7 \\
 & Deep-Multilingual-Punct & 65.6 & 56.5 & 63.8 \\
\midrule
\multirow{3}{*}{AMI} & Universal-2-TF & 77.5 & 69.7 & 82.7 \\
 & BadCode & 54.6 & 52.1 & 60.2 \\
 & Deep-Multilingual-Punct & 64.0 & 55.9 & 70.8 \\
\midrule
\multirow{3}{*}{Europarl} & Universal-2-TF & 83.0 & 73.2 & 83.6 \\
 & BadCode & 66.2 & 61.8 & 51.1 \\
 & Deep-Multilingual-Punct & 90.8 & 81.2 & 84.9 \\
\midrule
\multirow{3}{*}{\textbf{Average}} & Universal-2-TF & 81.9 & 71.4 & 83.5 \\
 & BadCode & 61.5 & 57.2 & 59.9 \\
 & Deep-Multilingual-Punct & 74.2 & 64.0 & 71.6 \\
\bottomrule
\end{tabular}
\end{table*}

Table \ref{tab:punctuation_results} compares Universal-2-TF's first-stage model with two open-source punctuation models. To directly evaluate punctuation classification accuracy, we used F1 scores across three categories: periods, commas, and question marks. BadCode implements a Transformer-based multi-headed prediction model to solve the PR and truecasing tasks\footnote{\url{https://github.com/1-800-BAD-CODE/punctuators}}, while Deep-Multilingual-Punct leverages a Transformer encoder with a single head to perform PR only \cite{guhr-EtAl:2021:fullstop}. Universal-2-TF outperforms both models on average in all punctuation categories, demonstrating its superior punctuation effectiveness. On the Europarl dataset, Deep-Multilingual-Punct performed the best. We believe this is because the model was trained on the Europarl dataset, avoiding penalties from the domain mismatch between the training and test sets, as well as potential differences in punctuation styles.

\begin{table*}[t]
\footnotesize
\centering
\caption{Truecasing F1 scores across different models and datasets.}
\label{tab:capitalization_results}
\vspace{-.5em}
\begin{tabular}{llccc}
\toprule
\multirow{2}{*}{\textbf{Dataset}} & \multirow{2}{*}{\textbf{Model}} & \multicolumn{3}{c}{\textbf{F1 score}} \\ \cmidrule(lr){3-5}

 & & $\mathtt{LOWER}$ & $\mathtt{ACRONYM}$ & $\mathtt{CAPITAL}$ \\
\midrule
\multirow{2}{*}{SummScreen} & Universal-2-TF & 97.5 & 95.8 & 88.7 \\
 & BadCode & 95.0 & 95.7 & 72.9 \\
\midrule
\multirow{2}{*}{DialogSum} & Universal-2-TF & 98.4 & 95.0 & 91.0 \\
 & BadCode & 96.6 & 97.9 & 79.7 \\
\midrule
\multirow{2}{*}{MeetingBank} & Universal-2-TF & 97.4 & 95.1 & 83.0 \\
 & BadCode & 96.0 & 93.8 & 73.8 \\
\midrule
\multirow{2}{*}{AMI} & Universal-2-TF & 97.7 & 95.4 & 79.7 \\
 & BadCode & 96.2 & 98.1 & 59.5 \\
\midrule
\multirow{2}{*}{Europarl} & Universal-2-TF & 98.3 & 95.2 & 86.1 \\
 & BadCode & 96.6 & 95.5 & 71.9 \\
\midrule
\multirow{2}{*}{\textbf{Average}} & Universal-2-TF & 97.9 & 95.3 & 85.7 \\
 & BadCode & 96.1 & 96.2 & 71.6 \\
\bottomrule
\end{tabular}
\end{table*}

Table \ref{tab:capitalization_results} compares the performance of Universal-2-TF's first-stage model and BadCode in terms of F1 scores across three truecasing categories: lower-case, acronyms, and capital-case. On average, both models performed similarly in lower-case and acronym word prediction, with BadCode being sometimes even better on acronym prediction. This may be due to the character-level prediction approach: in presence of consistent uppercase pattern, the model seems to succeed at capturing the casing global consensus. However, when the model must decide the case of a single letter within a word, the lack of consistent patterns makes it more difficult to establish a strong consensus for the correct casing, potentially increasing the likelihood of incorrect predictions or hallucinations.

\begin{table}[t]
    \footnotesize
    \centering
    \caption{I-WER comparison between NeMo and Universal-2-TF across various datasets.}
    \label{tab:itn_results}
    \vspace{-.5em}
    \begin{tabular}{llcc}
        \toprule
        \multirow{2}{*}{\textbf{Source}} & \multirow{2}{*}{\textbf{Dataset}} & \multicolumn{2}{c}{\textbf{Model}} \\ \cmidrule(lr){3-4}
         & & \textbf{NeMo} & \textbf{Universal-2-TF} \\
        \midrule
        \multirow{5}{*}{Public} & SummScreen & 78.6 & 11.5 \\
        & DialogSum & 62.3 & 29.7 \\
        & Europarl & 64.3 & 24.3 \\
        & MeetingBank & 49.0 & 18.9 \\
        & AMI & 29.7 & 13.1 \\
        \midrule
        \multirow{7}{*}{Private}
        & Credit Cards & 72.6 & 19.0 \\
        & Emails & 63.1 & 19.7 \\
        & Phones & 56.0 & 18.4 \\
        & PO Addresses & 49.8 & 14.2 \\
        & Web Addresses & 55.2 & 28.1 \\
        & SSN & 57.2 & 18.8 \\
        \midrule
        \multicolumn{2}{c}{\textbf{Average}} & 57.7 & 20.1 \\
        \bottomrule
    \end{tabular}
\end{table}

\begin{table}[t]
\footnotesize
\centering
\caption{ITN accuracy comparison using GTN dataset.}
\label{tab:gtn_comparison}
\vspace{-.5em}
\begin{tabular}{l c}
\toprule
Model                  & WER (\%) \\
\midrule
Nemo ITN~\cite{DBLP:journals/corr/abs-2104-05055}              & 12.7     \\
Thutmose (BERT)~\cite{antonova2022thutmose}       & 3.7      \\
Thutmose (d-BERT)~\cite{antonova2022thutmose}     & 3.7     \\
Neural ITN~\cite{9414912}            & 0.9      \\
Universal-2-TF        & 2.3     \\
\bottomrule
\end{tabular}
\end{table}

Table \ref{tab:itn_results} compares Universal-2-TF and NeMo's ITN performance using I-WER as the evaluation metric. In addition to the five publicly available datasets mentioned earlier, we used six internal test sets. These focus specifically on distinct ITN entities: credit card numbers, email addresses, phone numbers, postal addresses, URLs, and SSNs. Each test set contains 2,000 samples. Universal-2-TF significantly outperformed NeMo's ITN, which uses a WFST-based approach, demonstrating the advantage of a neural network-based method. 

Table \ref{tab:gtn_comparison} shows another ITN evaluation results obtained using the Google Text Normalization Challenge data~\cite{sproat2017rnnapproachestextnormalization}. Universal-2-TF outperformed all models, except for Neural ITN. This evaluation dataset contains extremely short-form data while Universal-2-TF is trained on to process longer segments of text. In real-world use cases, longer transcripts are expected, which contrasts with this dataset. Nonetheless, Universal-2-TF demonstrated competitive performance on short form texts as well.

\subsection{Examples}
\label{sec:examples}

Figure \ref{fig:examples} shows several TF examples comparing the proposed Universal-2-TF model with our previous model for different ITN entity categories. The proposed model demonstrates consistent formatting. It can be seen that Universal-2-TF generalizes well to different types of credit card numbers (15- and 16-digit numbers) and phones numbers on which Universal-1-TF struggles. Universal-2-TF also consistently formats email, URLs, and postal addresses correctly. It is worth noting that this improvement has been achieved with lower processing time, as shown in Table \ref{tab:e2e_results}.

\section{Conclusion}
\label{sec:conclusion}

In this paper, we proposed an all-neural TF (Text Formatting) model that achieves high accuracy across all TF aspects, including PR (punctuation restoration), truecasing, and ITN (inverse text normalization), as measured on a wide range of test sets. The model, used in our Universal-2 ASR system, consists of a multi-objective token classification model and a seq2seq model for ITN and mixed-case word conversion. This architecture ensures robustness across diverse text domains while maintaining efficient inference. Subjective evaluations of TF quality, conducted by external vendors, also confirm the superiority of the proposed model over its predecessor.

We emphasized the importance of building a comprehensive TF system and measuring its performance holistically, as highlighted in our previous work on ASR~\cite{ramirez2024anatomyindustrialscalemultilingual}. We believe the significance of such a systemic approach is growing as ASR is already widely employed in practical applications and is being increasingly adopted across a broader range of domains.

While the proposed model achieves impressive TF performance, it has some limitations. Currently, it can predict only post-sentence punctuation marks, which is inadequate for certain non-English languages, such as Spanish, which use pre-sentence punctuations like inverted question marks. This limitation could be addressed by adding an additional classification head to the first-stage model. Another potential improvement is to leverage acoustic input, which would enhance PR and enable the prediction of exclamation marks. However, real-world use cases operating at scale often demand very low processing latency and cost. Therefore, any solution must introduce minimal additional computational complexity. Future work should address these challenges.

\begin{figure*}[t]
\footnotesize
\mbox{}~\textbf{16-Digit Credit Card Numbers}\\
\vspace*{.5em}
\fbox{%
\begin{minipage}{\linewidth}
\textbf{Input:} \texttt{i'm so glad i finally reached someone my amazon order is a mess the confirmation email says three thousand four hundred and fifty six seven thousand eight hundred and ninety twelve thirty four five thousand six hundred and seventy eight was charged but i don't recognize that number
}\\
\textbf{Universal-2-TF:} \texttt{I'm so glad I finally reached someone. My Amazon order is a mess. The confirmation email says 3456-7890-1234-5678 was charged, but I don't recognize that number.
}\\
\textbf{Universal-1-TF:} \texttt{I'm so glad I finally reached someone. My Amazon order is a mess. The confirmation email says 3450 six. Seven thousand 890 twelve. Thirty four. Five thousand 678 was charged, but I don't recognize that number.
}
\end{minipage}%
}\\
\vspace*{0.7em}\\
\mbox{}~\textbf{15-Digit Credit Card Numbers}\\
\vspace*{.5em}
\fbox{%
\begin{minipage}{\linewidth}
\textbf{Input:} \texttt{they even provided me with a five seven zero eight two nine four six three two one zero two five eight four credit card to cover any incidental expenses i might have while on the trip
}\\
\textbf{Universal-2-TF:} \texttt{They even provided me with a 5782-946321-02584 credit card to cover any incidental expenses I might have while on the trip.
}\\
\textbf{Universal-1-TF:} \texttt{They even provided me with a 5708 two 9463-210-2584 credit card to cover any incidental expenses I might have while on the trip.
}
\end{minipage}%
}\\
\vspace*{0.7em}\\
\mbox{}~\textbf{Phone Numbers}\\
\vspace*{.5em}
\fbox{%
\begin{minipage}{\linewidth}
\textbf{Input:} \texttt{luckily she was smart enough to hang up and call the real ssa whose number is one eight zero zero seven seven two one two one three to report the scam}\\
\textbf{Universal-2-TF:} \texttt{Luckily, she was smart enough to hang up and call the real SSA, whose number is 1-800-772-1213 to report the scam.}\\
\textbf{Universal-1-TF:} \texttt{Luckily, she was smart enough to hang up and call the real SSA, whose number is 1800 770 212 13 to report the scam.}
\end{minipage}%
}\\
\vspace*{0.7em}\\
\mbox{}~\textbf{Websites and Emails}\\
\vspace*{.5em}
\fbox{%
\begin{minipage}{\linewidth}
\textbf{Input:} \texttt{if you're interested in learning more about our work you can check out our website at ai two one dot labs dot com or reach out to us at info at ai two one dot labs dot com
}\\
\textbf{Universal-2-TF:} \texttt{If you're interested in learning more about our work, you can check out our website at ai21.labs.com or reach out to us at info@ai21.labs.com.
}\\
\textbf{Universal-1-TF:} \texttt{If you're interested in learning more about our work, you can check out our website at AI two one dot labs.com or reach out to us at info at AI two one dot labs dot com.
}
\end{minipage}%
}\\
\vspace*{0.7em}\\
\mbox{}~\textbf{Postal Addresses}\\
\vspace*{.5em}
\fbox{%
\begin{minipage}{\linewidth}
\textbf{Input:} \texttt{the package is coming from amazon logistics originating at their fulfillment center on one twenty three industrial drive in newark new jersey zero seven one zero five
}\\
\textbf{Universal-2-TF:} \texttt{The package is coming from Amazon Logistics, originating at their fulfillment center on 123 Industrial Drive in Newark, New Jersey, 07105.
}\\
\textbf{Universal-1-TF:} \texttt{The package is coming from Amazon Logistics, originating at their fulfillment center on 123 industrial drive in Newark, New Jersey. Zero seven 10 five.
}
\end{minipage}%
}
\caption{Text formatting examples comparing Universal-2-TF (proposed model) with Universal-1-TF (previous model).}
\label{fig:examples}
\end{figure*}

\section*{Acknowledgements}

We thank Ahmed Etefy for developing the infrastructure required to process training data at scale; Rajpreet Thethy for support with subjective evaluations; Mimi Chiang and Aleks Mitov for engineering support; Michael Liang for assistance with quantitative evaluations; and other Research team members for discussions.

\bibliography{acl2023}

\begin{thebibliography}{33}
\expandafter\ifx\csname natexlab\endcsname\relax\def\natexlab#1{#1}\fi

\bibitem[{Antonova et~al.(2022)Antonova, Bakhturina, and
  Ginsburg}]{antonova2022thutmose}
Alexandra Antonova, Evelina Bakhturina, and Boris Ginsburg. 2022.
\newblock \href {https://doi.org/10.21437/INTERSPEECH.2022-10864} {Thutmose
  tagger: Single-pass neural model for inverse text normalization}.
\newblock In \emph{23rd Annual Conference of the International Speech
  Communication Association, Interspeech 2022, Incheon, Korea, September 18-22,
  2022}, pages 550--554. {ISCA}.

\bibitem[{AssemblyAI(2024)}]{assemblyai2024punctuation}
AssemblyAI. 2024.
\newblock \href
  {https://www.assemblyai.com/blog/introducing-our-new-punctuation-restoration-and-truecasing-models/}
  {Introducing our new punctuation restoration and truecasing models}.

\bibitem[{Carletta et~al.(2006)Carletta, Ashby, Bourban, Flynn, Guillemot,
  Hain, Kadlec, Karaiskos, Kraaij, Kronenthal, Lathoud, Lincoln, Lisowska,
  McCowan, Post, Reidsma, and Wellner}]{AMICorpus}
Jean Carletta, Simone Ashby, Sebastien Bourban, Mike Flynn, Mael Guillemot,
  Thomas Hain, Jaroslav Kadlec, Vasilis Karaiskos, Wessel Kraaij, Melissa
  Kronenthal, Guillaume Lathoud, Mike Lincoln, Agnes Lisowska, Iain McCowan,
  Wilfried Post, Dennis Reidsma, and Pierre Wellner. 2006.
\newblock The ami meeting corpus: A pre-announcement.
\newblock In \emph{Machine Learning for Multimodal Interaction}, pages 28--39,
  Berlin, Heidelberg. Springer Berlin Heidelberg.

\bibitem[{Chen et~al.(2022)Chen, Chu, Wiseman, and
  Gimpel}]{chen2022summscreendatasetabstractivescreenplay}
Mingda Chen, Zewei Chu, Sam Wiseman, and Kevin Gimpel. 2022.
\newblock \href {https://doi.org/10.18653/v1/2022.acl-long.589}
  {{S}umm{S}creen: A dataset for abstractive screenplay summarization}.
\newblock In \emph{Proceedings of the 60th Annual Meeting of the Association
  for Computational Linguistics (Volume 1: Long Papers)}, pages 8602--8615,
  Dublin, Ireland. Association for Computational Linguistics.

\bibitem[{Chen et~al.(2021)Chen, Liu, Chen, and
  Zhang}]{chen2021dialogsumreallifescenariodialogue}
Yulong Chen, Yang Liu, Liang Chen, and Yue Zhang. 2021.
\newblock \href {https://doi.org/10.18653/v1/2021.findings-acl.449}
  {{D}ialog{S}um: {A} real-life scenario dialogue summarization dataset}.
\newblock In \emph{Findings of the Association for Computational Linguistics:
  ACL-IJCNLP 2021}, pages 5062--5074, Online. Association for Computational
  Linguistics.

\bibitem[{Courtland et~al.(2020)Courtland, Faulkner, and
  McElvain}]{courtland2020efficient}
Maury Courtland, Adam Faulkner, and Gayle McElvain. 2020.
\newblock \href {https://doi.org/10.18653/v1/2020.iwslt-1.33} {Efficient
  automatic punctuation restoration using bidirectional transformers with
  robust inference}.
\newblock In \emph{Proceedings of the 17th International Conference on Spoken
  Language Translation}, pages 272--279, Online. Association for Computational
  Linguistics.

\bibitem[{Devlin et~al.(2019)Devlin, Chang, Lee, and
  Toutanova}]{devlin-etal-2018-bert}
Jacob Devlin, Ming-Wei Chang, Kenton Lee, and Kristina Toutanova. 2019.
\newblock \href {https://doi.org/10.18653/v1/N19-1423} {{BERT}: Pre-training of
  deep bidirectional transformers for language understanding}.
\newblock In \emph{Proceedings of the 2019 Conference of the North {A}merican
  Chapter of the Association for Computational Linguistics: Human Language
  Technologies, Volume 1 (Long and Short Papers)}, pages 4171--4186,
  Minneapolis, Minnesota. Association for Computational Linguistics.

\bibitem[{Gaur et~al.(2023)Gaur, Kibre, Xue, Shu, Wang, Alphanso, Li, and
  Gong}]{10022543}
Yashesh Gaur, Nick Kibre, Jian Xue, Kangyuan Shu, Yuhui Wang, Issac Alphanso,
  Jinyu Li, and Yifan Gong. 2023.
\newblock \href {https://doi.org/10.1109/SLT54892.2023.10022543} {Streaming,
  fast and accurate on-device inverse text normalization for automatic speech
  recognition}.
\newblock In \emph{2022 IEEE Spoken Language Technology Workshop (SLT)}, pages
  237--244.

\bibitem[{Guerreiro et~al.(2021)Guerreiro, Rei, and
  Batista}]{guerreiro2021towards}
Nuno~Miguel Guerreiro, Ricardo Rei, and Fernando Batista. 2021.
\newblock \href {https://doi.org/https://doi.org/10.1016/j.eswa.2021.115740}
  {Towards better subtitles: A multilingual approach for punctuation
  restoration of speech transcripts}.
\newblock \emph{Expert Systems with Applications}, 186:115740.

\bibitem[{Guhr et~al.(2021)Guhr, Schumann, Bahrmann, and
  Böhme}]{guhr-EtAl:2021:fullstop}
Oliver Guhr, Anne-Kathrin Schumann, Frank Bahrmann, and Hans~Joachim Böhme.
  2021.
\newblock \href {http://ceur-ws.org/Vol-2957/sepp_paper4.pdf} {Fullstop:
  Multilingual deep models for punctuation prediction}.
\newblock In \emph{Proceedings of the Swiss Text Analytics Conference 2021},
  Winterthur, Switzerland. CEUR Workshop Proceedings.

\bibitem[{Hu et~al.(2023)Hu, Ganter, Deilamsalehy, Dernoncourt, Foroosh, and
  Liu}]{hu-etal-2023-meetingbank}
Yebowen Hu, Timothy Ganter, Hanieh Deilamsalehy, Franck Dernoncourt, Hassan
  Foroosh, and Fei Liu. 2023.
\newblock \href {https://doi.org/10.18653/v1/2023.acl-long.906}
  {{M}eeting{B}ank: A benchmark dataset for meeting summarization}.
\newblock In \emph{Proceedings of the 61st Annual Meeting of the Association
  for Computational Linguistics (Volume 1: Long Papers)}, pages 16409--16423,
  Toronto, Canada. Association for Computational Linguistics.

\bibitem[{Koehn(2005)}]{koehn-2005-europarl}
Philipp Koehn. 2005.
\newblock \href {https://aclanthology.org/2005.mtsummit-papers.11/}
  {{E}uroparl: A parallel corpus for statistical machine translation}.
\newblock In \emph{Proceedings of Machine Translation Summit X: Papers}, pages
  79--86, Phuket, Thailand.

\bibitem[{Lewis et~al.(2020)Lewis, Liu, Goyal, Ghazvininejad, Mohamed, Levy,
  Stoyanov, and Zettlemoyer}]{lewis-etal-2020-bart}
Mike Lewis, Yinhan Liu, Naman Goyal, Marjan Ghazvininejad, Abdelrahman Mohamed,
  Omer Levy, Veselin Stoyanov, and Luke Zettlemoyer. 2020.
\newblock \href {https://doi.org/10.18653/v1/2020.acl-main.703} {{BART}:
  Denoising sequence-to-sequence pre-training for natural language generation,
  translation, and comprehension}.
\newblock In \emph{Proceedings of the 58th Annual Meeting of the Association
  for Computational Linguistics}, pages 7871--7880, Online. Association for
  Computational Linguistics.

\bibitem[{Lin and Wang(2020)}]{lin2020joint}
Binghuai Lin and Liyuan Wang. 2020.
\newblock \href {https://doi.org/10.21437/Interspeech.2020-1277} {Joint
  prediction of punctuation and disfluency in speech transcripts}.
\newblock In \emph{Interspeech 2020}, pages 716--720.

\bibitem[{Meister et~al.(2023)Meister, Novikov, Karpov, Bakhturina, Lavrukhin,
  and Ginsburg}]{meister2023librispeechpcbenchmarkevaluationpunctuation}
Aleksandr Meister, Matvei Novikov, Nikolay Karpov, Evelina Bakhturina, Vitaly
  Lavrukhin, and Boris Ginsburg. 2023.
\newblock \href {https://arxiv.org/abs/2310.02943} {Librispeech-pc: Benchmark
  for evaluation of punctuation and capitalization capabilities of end-to-end
  asr models}.
\newblock \emph{arXiv preprint arXiv:2310.02943}.

\bibitem[{Nguyen et~al.(2019)Nguyen, Nguyen, Nguyen, Phuong, Nguyen, Do, and
  Mai}]{nguyen2019fast}
Binh Nguyen, Vu~Bao~Hung Nguyen, Hien Nguyen, Pham~Ngoc Phuong, The-Loc Nguyen,
  Quoc~Truong Do, and Luong~Chi Mai. 2019.
\newblock \href {https://doi.org/10.1109/O-COCOSDA46868.2019.9041202} {Fast and
  accurate capitalization and punctuation for automatic speech recognition
  using transformer and chunk merging}.
\newblock In \emph{2019 22nd Conference of the Oriental COCOSDA International
  Committee for the Co-ordination and Standardisation of Speech Databases and
  Assessment Techniques (O-COCOSDA)}, pages 1--5.

\bibitem[{Nguyen et~al.(2023)Nguyen, Nguyen, Do, Luong, Waibel
  et~al.}]{nguyen2023adapitn}
Thai-Binh Nguyen, Quang~Minh Nguyen, Quoc~Truong Do, Chi~Mai Luong, Alexander
  Waibel, et~al. 2023.
\newblock \href {https://openreview.net/forum?id=VI9IWawOr3} {Adapitn: A fast,
  reliable, and dynamic adaptive inverse text normalization}.
\newblock In \emph{ICASSP 2023-2023 IEEE International Conference on Acoustics,
  Speech and Signal Processing (ICASSP)}, pages 1--5. IEEE.

\bibitem[{Pappagari et~al.(2021)Pappagari, Żelasko, Mikołajczyk, Pęzik, and
  Dehak}]{9687976}
Raghavendra Pappagari, Piotr Żelasko, Agnieszka Mikołajczyk, Piotr Pęzik,
  and Najim Dehak. 2021.
\newblock \href {https://doi.org/10.1109/ASRU51503.2021.9687976} {Joint
  prediction of truecasing and punctuation for conversational speech in
  low-resource scenarios}.
\newblock In \emph{2021 IEEE Automatic Speech Recognition and Understanding
  Workshop (ASRU)}, pages 1185--1191.

\bibitem[{Paul et~al.(2022)Paul, Pang, Chen, and Zhang}]{paul2022improving}
Debjyoti Paul, Yutong Pang, Szu-Jui Chen, and Xuedong Zhang. 2022.
\newblock \href
  {https://www.isca-archive.org/interspeech_2022/paul22_interspeech.html}
  {Improving data driven inverse text normalization using data augmentation and
  machine translation}.
\newblock In \emph{Interspeech 2022}, pages 5221--5222.

\bibitem[{Pusateri et~al.(2017)Pusateri, Ambati, Brooks, Platek, McAllaster,
  and Nagesha}]{pusateri2017mostly}
Ernest Pusateri, Bharat~Ram Ambati, Elizabeth Brooks, Ondrej Platek, Donald
  McAllaster, and Venki Nagesha. 2017.
\newblock \href {https://doi.org/10.21437/Interspeech.2017-1274} {A mostly
  data-driven approach to inverse text normalization}.
\newblock In \emph{Interspeech 2017}, pages 2784--2788.

\bibitem[{Radford et~al.(2023)Radford, Kim, Xu, Brockman, McLeavey, and
  Sutskever}]{radford2023robust}
Alec Radford, Jong~Wook Kim, Tao Xu, Greg Brockman, Christine McLeavey, and
  Ilya Sutskever. 2023.
\newblock \href {https://dl.acm.org/doi/10.5555/3618408.3619590} {Robust speech
  recognition via large-scale weak supervision}.
\newblock In \emph{Proceedings of the 40th International Conference on Machine
  Learning}, ICML'23. JMLR.org.

\bibitem[{Ramena et~al.(2020)Ramena, Nagaraju, Moharana, Prasanna~Mohanty, and
  Purre}]{ramena2020efficient}
Gopi Ramena, Divija Nagaraju, Sukumar Moharana, Debi Prasanna~Mohanty, and
  Naresh Purre. 2020.
\newblock \href {https://doi.org/10.1109/ICSC.2020.00035} {{ An Efficient
  Architecture for Predicting the Case of Characters using Sequence Models }}.
\newblock In \emph{2020 IEEE 14th International Conference on Semantic
  Computing (ICSC)}, pages 174--177, Los Alamitos, CA, USA. IEEE Computer
  Society.

\bibitem[{Ramirez et~al.(2024)Ramirez, Chkhetiani, Ehrenberg, McHardy, Botros,
  Khare, Vanzo, Peyash, Oexle, Liang, Sklyar, Fakhan, Etefy, McCrystal,
  Flamini, Donato, and
  Yoshioka}]{ramirez2024anatomyindustrialscalemultilingual}
Francis~McCann Ramirez, Luka Chkhetiani, Andrew Ehrenberg, Robert McHardy, Rami
  Botros, Yash Khare, Andrea Vanzo, Taufiquzzaman Peyash, Gabriel Oexle,
  Michael Liang, Ilya Sklyar, Enver Fakhan, Ahmed Etefy, Daniel McCrystal, Sam
  Flamini, Domenic Donato, and Takuya Yoshioka. 2024.
\newblock \href {http://arxiv.org/abs/2404.09841} {Anatomy of industrial scale
  multilingual asr}.
\newblock \emph{arXiv preprint arXiv:2404.09841}.

\bibitem[{Sproat and Jaitly(2017)}]{sproat2017rnnapproachestextnormalization}
Richard Sproat and Navdeep Jaitly. 2017.
\newblock \href {http://arxiv.org/abs/1611.00068} {Rnn approaches to text
  normalization: A challenge}.
\newblock \emph{arXiv preprint arXiv:1611.00068}.

\bibitem[{Sunkara et~al.(2020)Sunkara, Ronanki, Dixit, Bodapati, and
  Kirchhoff}]{sunkara-etal-2020-robust}
Monica Sunkara, Srikanth Ronanki, Kalpit Dixit, Sravan Bodapati, and Katrin
  Kirchhoff. 2020.
\newblock \href {https://doi.org/10.18653/v1/2020.nlpmc-1.8} {Robust prediction
  of punctuation and truecasing for medical {ASR}}.
\newblock In \emph{Proceedings of the First Workshop on Natural Language
  Processing for Medical Conversations}, pages 53--62, Online. Association for
  Computational Linguistics.

\bibitem[{Sunkara et~al.(2021)Sunkara, Shivade, Bodapati, and
  Kirchhoff}]{9414912}
Monica Sunkara, Chaitanya Shivade, Sravan Bodapati, and Katrin Kirchhoff. 2021.
\newblock \href {https://doi.org/10.1109/ICASSP39728.2021.9414912} {Neural
  inverse text normalization}.
\newblock In \emph{ICASSP 2021 - 2021 IEEE International Conference on
  Acoustics, Speech and Signal Processing (ICASSP)}, pages 7573--7577.

\bibitem[{Susanto et~al.(2016)Susanto, Chieu, and Lu}]{susanto2016learning}
Raymond~Hendy Susanto, Hai~Leong Chieu, and Wei Lu. 2016.
\newblock \href {https://doi.org/10.18653/v1/D16-1225} {Learning to capitalize
  with character-level recurrent neural networks: An empirical study}.
\newblock In \emph{Proceedings of the 2016 Conference on Empirical Methods in
  Natural Language Processing}, pages 2090--2095, Austin, Texas. Association
  for Computational Linguistics.

\bibitem[{Tan et~al.(2023)Tan, Behre, Kibre, Alphonso, and Chang}]{tan2023four}
Sharman Tan, Piyush Behre, Nick Kibre, Issac Alphonso, and Shuangyu Chang.
  2023.
\newblock \href {https://doi.org/10.1109/SLT54892.2023.10023257} {Four-in-one:
  a joint approach to inverse text normalization, punctuation, capitalization,
  and disfluency for automatic speech recognition}.
\newblock In \emph{2022 IEEE Spoken Language Technology Workshop (SLT)}, pages
  677--684.

\bibitem[{Tilk and Alumäe(2015)}]{tilk2015lstm}
Ottokar Tilk and Tanel Alumäe. 2015.
\newblock \href {https://doi.org/10.21437/Interspeech.2015-240} {Lstm for
  punctuation restoration in speech transcripts}.
\newblock In \emph{Interspeech 2015}, pages 683--687.

\bibitem[{Zhang et~al.(2022)Zhang, Cheng, Kumar, Huang, Chen, and
  Mathews}]{zhang2022capitalization}
Hao Zhang, You-Chi Cheng, Shankar Kumar, W.~Ronny Huang, Mingqing Chen, and
  Rajiv Mathews. 2022.
\newblock \href {https://doi.org/10.1109/ICASSP43922.2022.9746492}
  {Capitalization normalization for language modeling with an accurate and
  efficient hierarchical rnn model}.
\newblock In \emph{ICASSP 2022 - 2022 IEEE International Conference on
  Acoustics, Speech and Signal Processing (ICASSP)}, pages 6097--6101.

\bibitem[{Zhang et~al.(2021)Zhang, Bakhturina, Gorman, and
  Ginsburg}]{DBLP:journals/corr/abs-2104-05055}
Yang Zhang, Evelina Bakhturina, Kyle Gorman, and Boris Ginsburg. 2021.
\newblock \href {http://arxiv.org/abs/2104.05055} {Nemo inverse text
  normalization: From development to production}.
\newblock \emph{CoRR}, abs/2104.05055.

\bibitem[{Zhu et~al.(2024)Zhu, Chang, Gardiner, Rossouw, and
  Robertson}]{zhu2024resolving}
Xiliang Zhu, Chia-Tien Chang, Shayna Gardiner, David Rossouw, and Jonas
  Robertson. 2024.
\newblock \href {https://aclanthology.org/2024.unimplicit-1.3/} {Resolving
  transcription ambiguity in {S}panish: A hybrid acoustic-lexical system for
  punctuation restoration}.
\newblock In \emph{Proceedings of the Third Workshop on Understanding Implicit
  and Underspecified Language}, pages 33--41, Malta. Association for
  Computational Linguistics.

\bibitem[{Żelasko et~al.(2018)Żelasko, Szymański, Mizgajski, Szymczak,
  Carmiel, and Dehak}]{zelasko2018punctuation}
Piotr Żelasko, Piotr Szymański, Jan Mizgajski, Adrian Szymczak, Yishay
  Carmiel, and Najim Dehak. 2018.
\newblock \href {https://doi.org/10.21437/Interspeech.2018-1096} {Punctuation
  prediction model for conversational speech}.
\newblock In \emph{Interspeech 2018}, pages 2633--2637.

\end{thebibliography}
\bibliographystyle{acl_natbib}

\appendix

\section{Details of Data Cleaning}
\label{sec:appendix}

After collecting the data and examining random samples, we identified numerous artifacts in the datasets that introduced noise into the training process. The artifacts range from random characters such as \#, \&, <, >, and \_, to additional punctuation marks like !, ;, :, [], and (), which our models have not yet been trained to predict. To clean the data and eliminate the noise, we set up a data processing pipeline to filter out noisy samples and clean the remaining ones.

The main heuristics we apply to clean our training data were as follows. These heuristics were developed based on thorough examinations of the training data.
\begin{itemize}
\item Remove parentheses, square brackets, and curly brackets along with the words within them.
\item Remove emojis, HTML tags, and any instances of speaker labels.
\item Remove any period, question mark, or comma at the start of the text if followed by a space.
\item Replace multiple consecutive punctuation marks with a single instance. Since we do not support exclamation marks yet, replace exclamation marks with periods.
\item Replace instances of multiple consecutive spaces and spaces before end-of-sentence punctuation.
\item Capitalize the first word of each sentence and uniformly standardize variations of ``Mr'', ``Dr'', ``Prof'' and ``ok'' to ``Mr.'', ``Dr.'', ``Prof.'', and ``OK'' respectively.
\item Loop through a list of common filler words such as ``Well'', ``Umm'', ``Mmm'', and convert them to lowercase if they appear in the middle of a sentence.
\item Convert capital letters following ellipses (...) to lowercase.
\item Convert ``You'' or ``Your'' to lowercase ``you'' or ``your'' if not preceded by a period or question mark.
\item Replace symbols such as `¿', `¡', `©', `®', `¦', `¬', `\_', etc.
\end{itemize}

In addition to applying these heuristics to remove irregularities in the transcripts, we also filtered out training samples that had unusually high or low numbers of uppercase letters or punctuation marks relative to the sentence length.

\end{document}